%% file: paper.tex
\documentclass[]{damo}
\usepackage{wrapfig}
\usepackage{tabularx}
\usepackage{textcomp}
\usepackage{stfloats}
\usepackage{url}
\usepackage{verbatim}
\usepackage{graphicx}
\usepackage{titlesec}
\usepackage{tocloft}
\usepackage{adjustbox}
\usepackage{multirow}
\usepackage{pifont}
\usepackage{tikz}
\usepackage{comment}
\usepackage{amsmath,amssymb} % define this before the line numbering.
\usepackage{colortbl}  %彩色表格需要加载的宏包
\usepackage{color}
\usepackage{booktabs} 
\usepackage{hyperref}
\usepackage{graphicx}    % 用于插入图片
\usepackage{subcaption} 
\usepackage{booktabs}
\usepackage{amsmath} % 数学公式支持
\usepackage{multirow} % 表格多行合并
\usepackage{booktabs} % 表格更好的线条
\usepackage{subcaption} % 支持 subtable 和 subfigure

\RequirePackage{xspace}
\makeatletter
\DeclareRobustCommand\onedot{\futurelet\@let@token\@onedot}
\def\@onedot{\ifx\@let@token.\else.\null\fi\xspace}

\makeatother

\definecolor{adptorange}{RGB}{248, 205, 172}
\definecolor{cmpblue}{RGB}{189, 215, 238}
\definecolor{cmpblue}{RGB}{189, 215, 238}

\definecolor{our_red}{RGB}{232,157,160}
\definecolor{our_blue}{RGB}{136,206,230}
\definecolor{our_orange}{RGB}{246,200,168}
\definecolor{our_green}{RGB}{178,211,164}

\definecolor{attn_code0}{RGB}{247,215,200}
\definecolor{attn_code1}{RGB}{238,169,139}
\definecolor{mlp_code0}{RGB}{204,201,221}
\definecolor{mlp_code1}{RGB}{102,95,153}

\definecolor{token_blue}{RGB}{84, 120, 140}
\usepackage{pifont}       % \ding{xx}
\usepackage{bbding}       % \Checkmark,\XSolid,... (需要和pifont宏包共同使用)
\usepackage{fontawesome}
\usepackage{xspace}

\usepackage{float}

\newlength\savewidth

\newcolumntype{x}[1]{>{\centering\arraybackslash}p{#1pt}}
\newcolumntype{y}[1]{>{\raggedright\arraybackslash}p{#1pt}}
\newcolumntype{z}[1]{>{\raggedleft\arraybackslash}p{#1pt}}

\renewcommand{\paragraph}[1]{\vspace{1mm}\noindent\textbf{#1}}
\usepackage{colortbl}
\usepackage{xcolor}
\usepackage{wrapfig}

\renewcommand{\paragraph}[1]{\vspace{1.25mm}\noindent\textbf{#1}}

\usepackage{algorithm}
\usepackage{listings}

\definecolor{codeblue}{rgb}{0.25, 0.5, 0.5}
\definecolor{codekw}{rgb}{0.35, 0.35, 0.75}
\lstdefinestyle{Pytorch}{
    language = Python,
    backgroundcolor = \color{white},
    basicstyle = \fontsize{9pt}{8pt}\selectfont\ttfamily\bfseries,
    columns = fullflexible,
    aboveskip=1pt,
    belowskip=1pt,
    breaklines = true,
    captionpos = b,
    commentstyle = \color{codeblue},
    keywordstyle = \color{codekw},
}

\definecolor{green}{HTML}{009000}
\definecolor{red}{HTML}{ea4335}

\title{JCo-MVTON: Jointly Controllable Multi-Modal Diffusion Transformer for Mask-Free Virtual Try-on}

\author[* 1,3]{Aowen Wang}
\author[* 1]{Wei Li}
\author[\dagger 1, 2]{Hao Luo}
\author[1]{Mengxing Ao}
\author[3]{Chenyu Zhu}
\author[1,3]{Xinyang Li}
\author[1]{Fan Wang}

\affiliation[1]{DAMO Academy, Alibaba Group \\}
\affiliation[2]{Hupan Lab \\}
\affiliation[3]{Zhejiang University}
\contribution[*]{Equal contribution}
\contribution[\dagger]{Corresponding author}

\input{sec/0_abstract}

\date{\today}

\begin{document}
\thispagestyle{firstheader}
\maketitle
\pagestyle{empty}

\input{sec/1_introduction}

\input{sec/2_relatedwork}
\input{sec/3_method}

\input{sec/4_experiment}

\input{sec/5_conclusion}

\bibliographystyle{assets/plainnat}
\bibliography{paper}

\end{document}

%% file: sec/0_abstract.tex
\abstract{
Virtual try-on systems have long struggled with rigid dependencies on human body masks, limited fine-grained control over garment attributes, and poor generalization to in-the-wild scenarios. In this paper, we propose \textbf{JCo-MVTON} (\textbf{J}ointly \textbf{C}ontrollable Multi-Modal Diffusion Transformer for \textbf{M}ask-Free \textbf{V}irtual \textbf{T}ry-\textbf{O}n), a novel framework that simultaneously addresses these challenges by unifying diffusion-based generation with multi-modal condition fusion. Our architecture leverages a Multi-Modal Diffusion Transformer (MM-DiT) backbone to integrate diverse control signals—including reference image and garment image—directly into the denoising process. Our key architectural innovation involves conditioning the MM-DiT by fusing reference and garment information into its self-attention layers through dedicated conditional pathways. This integration is augmented by critical refinements to the positional encodings and attention masks, which specialize the multi-conditional MM-DiT for the nuanced requirements of the virtual try-on task and yield superior results. On the data curation front, we introduce a bi-directional generation strategy to construct our training set. This approach leverages two complementary pathways: first, a mask-based model is employed to generate a substantial volume of reference images; second, a Try-Off model, sharing an identical architecture and trained via self-supervision, synthesizes the corresponding garment data. Furthermore, the entire generated corpus undergoes a meticulous manual screening process, allowing us to iteratively improve the quality and fidelity of the training data. Our method achieves state-of-the-art performance on public benchmarks (e.g., DressCode) and outperforms commercial systems in real-world scenarios. The online API demo is available at \href{https://market.aliyun.com/apimarket/detail/cmapi00067129?spm=5176.shop.result.2.6e323934OAW8XR&innerSource=search}{Alibaba Cloud Marketplace}.

\smallskip
\noindent\textbf{Project page and code}: \href{https://github.com/damo-cv/JCo-MVTON}{https://github.com/damo-cv/JCo-MVTON}
}

%% file: sec/1_introduction.tex
\section{Introduction} \label{sec:introduction}
Virtual try-on (VTON) systems aim to synthesize a realistic image of a person wearing a new garment while preserving the person’s original pose, body shape, and appearance as faithfully as possible. This technology has become increasingly important for online fashion and retail: by enabling users to “virtually try on” clothing and accessories prior to purchasing them, VTON addresses common uncertainties about fit, proportion, and style compatibility, thereby enhancing customer satisfaction, increasing user engagement, and reducing return rates in e-commerce \citep{han2018viton, islam2023ecommerce}. As online shopping continues to grow, the lack of physical interaction with garments remains a key limitation; VTON bridges this gap by providing an immersive and personalized shopping experience that mimics in-store fitting rooms.

\begin{figure}[h!] 
    \centering % 图片居中
    \includegraphics[width=0.9\textwidth]{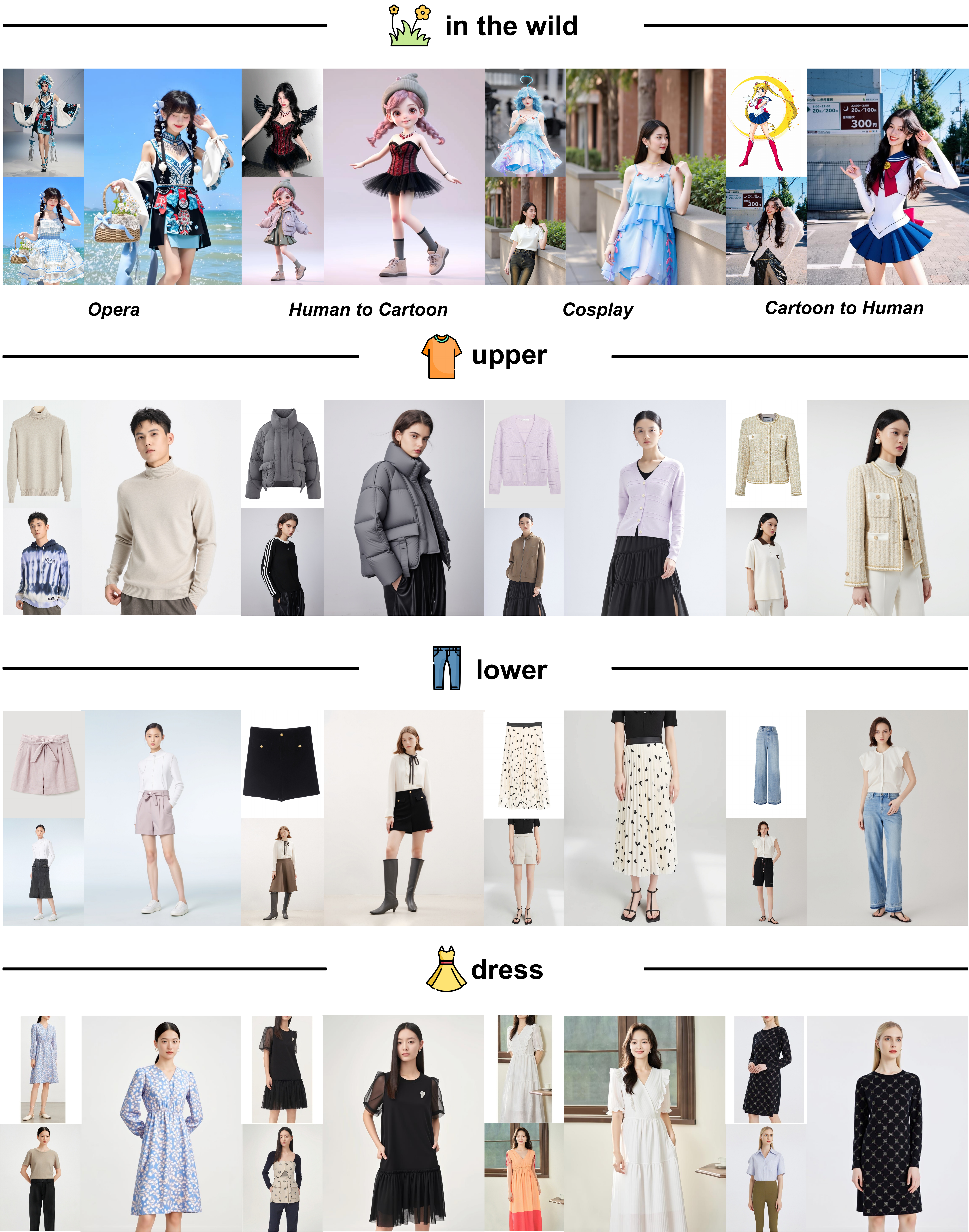}
    \caption{Our JCo-MVTON model for virtual try-on, a Multi-Modal Diffusion Transformer (MM-DiT) framework featuring an efficient data construction method and a specialized attention mechanism to achieve state-of-the-art results.} % 添加图片标题
    \label{fig:my_example} % 添加标签，用于在文中引用
\end{figure}

Early VTON methods generally followed a multi-stage pipeline based on generative adversarial networks (GANs) and warping modules. These approaches warp the garment onto the target body (e.g., via thin-plate splines or learned flow fields) and then use a GAN-based inpainting network to blend it with the person \citep{han2018viton, chen2024diffusion_vton}. However, GAN-based models often struggle to produce high-quality results under challenging poses and complex backgrounds. Recent surveys report that diffusion models have begun to surpass GANs in both fidelity and diversity for image synthesis tasks, including VTON \citep{lee2023survey, chen2024diffusion_vton}.

Broadly, diffusion-based VTON methods fall into two camps: mask-based and mask-free approaches \citep{liu2024maskfree, wang2024boow}. In mask-based pipelines, one first generates a binary human-parsing mask to occlude the subject’s original apparel region, then warps the target garment to align with that mask, and finally applies a diffusion (or GAN) model to inpaint the occluded area \citep{han2018viton, wang2024boow}. Although this formulation can yield finely detailed alignments, it entails two key drawbacks. First, the multi-stage nature of segmentation, warping, and inpainting makes the pipeline brittle: errors in mask prediction or garment alignment tend to cascade and compromise the final output. Second, reliance on hard masks limits flexibility, since occluding regions may discard critical context (e.g., hands, background elements) and demands highly accurate human parsing—a requirement that is often difficult to satisfy in real-world settings \citep{wang2024boow}.

The inpainting‐based paradigm benefits from a well‐defined, modular workflow—human parsing yields a precise clothing mask, which guides texture‐to‐body correspondence and simplifies subsequent inpainting stages \citep{han2018vitonimagebasedvirtualtryon, wan2025mfvitonhighfidelitymaskfreevirtual}. Nevertheless, it is prone to cascading failures: parsing errors or complex occlusions can produce visible artifacts (e.g., misplaced hair or arms, clothing “leakage”), while masking disrupts the original image’s spatial and lighting cues, degrading garment detail and background coherence \citep{jiang2024fitditadvancingauthenticgarment, atef2025efficientvitonefficientvirtualtryon, choi2024improvingdiffusionmodelsauthentic}.
By contrast, mask‐free methods dispense with explicit segmentation, thereby avoiding mask‐induced artifacts and preserving the scene’s innate geometry and illumination. During inference, only the person and garment images are required—no separate parsing, warping, or fusion steps—resulting in a more streamlined and robust pipeline \citep{zhang2024boowvtonboostinginthewildvirtual, niu2024pfdm, chang2025pemfvtopointenhancedvideovirtual}. Consequently, mask‐free diffusion models have emerged as the dominant paradigm for high‐fidelity virtual try-on.

However, the mask-free paradigm\citep{guo2025any2anytryonleveragingadaptiveposition,yang2024d4vtondynamicsemanticsdisentangling} also suffers from several drawbacks. First, mask-free methods entirely forgo the use of an explicit mask input for the target region, relying solely on the garment and model images; as a result, the network must infer all aspects of the clothing rendering automatically, which prevents precise, localized refinement along garment boundaries or in fine‐detail regions. A second challenge lies in the construction of triplet datasets\citep{sun2025dsvtonhighqualityvirtualtryon}: training in a mask-free manner requires simultaneous access to three aligned images: the original model, the isolated garment, and the post-try-on result. Compared to traditional mask-based approaches, this greatly complicates dataset assembly. Indeed, mainstream benchmarks (e.g., VITON\citep{han2018viton}, DressCode\citep{morelli2022dresscodehighresolutionmulticategory}) are generally released only after mask-based preprocessing and provide merely segmentation masks, without furnishing complete, pixel-level ground truth for all three elements.

To address the limitations inherent in mask-free approaches, we propose the Jointly Controllable Multi-Modal Diffusion Transformer for Mask-Free Virtual Try-On (JCo-MVTON). First, we introduce a cyclic data-preparation pipeline to overcome the scarcity of mask-free triplet datasets: by leveraging Try-Off \citep{velioglu2024tryoffdiffvirtualtryoffhighfidelitygarment}, mask-based Try-on, and IC-LoRA\citep{huang2024context} techniques, we generate a large pool of candidate triplets, which are then manually filtered to yield high-quality training samples. Through successive iterations of model refinement and data augmentation, this pipeline becomes self-sustaining. Moreover, our model employs a multi-condition MM-DiT mechanism \citep{esser2024scalingrectifiedflowtransformers} to achieve more stable try-on results; compared to the previous state-of-the-art ReferenceNet\citep{chen2024wearanywaymanipulablevirtualtryon}, MM-DiT significantly raises both the upper bound of generation quality and the consistency with the original garment. Finally, we further investigate the integration of positional encoding and a masked-attention mechanism to mitigate inter-branch interference, thereby enhancing the fidelity and alignment of generated outputs with the source model images.

Overall, our method achieves significant success on the virtual try-on task. We establish a sustainable, novel paradigm for generating mask-free training data that co-evolves with our model through continuous self-iteration and optimization, thereby cyclically improving both model performance and data quality. Furthermore, our model itself not only attains state-of-the-art results on benchmarks such as VITON-HD, but also demonstrates exceptional performance across diverse scenarios: varying garment types (tops and bottoms), styles (anime, real-world apparel, avant-garde costumes), and application contexts (B2B and B2C). This versatility evidences its high stability and image fidelity. In our human evaluation against existing commercial solutions, our approach markedly outperforms competitors on all assessed metrics—including Image Authenticity, Image Clarity, Silhouette Consistency, Detail Consistency, and Overall Harmony—underscoring its superiority in practical deployment.

\section{Human Evaluation}

In order to evaluate the performance of different models in virtual try-on tasks, we designed a human evaluation framework based on five key metrics: \textit{Image Authenticity}, \textit{Image Clarity}, \textit{Silhouette Consistency}, \textit{Detail Consistency}, and \textit{Overall Harmony}. The definitions of these metrics are as follows:
\begin{figure}[h!] 
    \centering % 图片居中
    \includegraphics[width=0.9\textwidth]{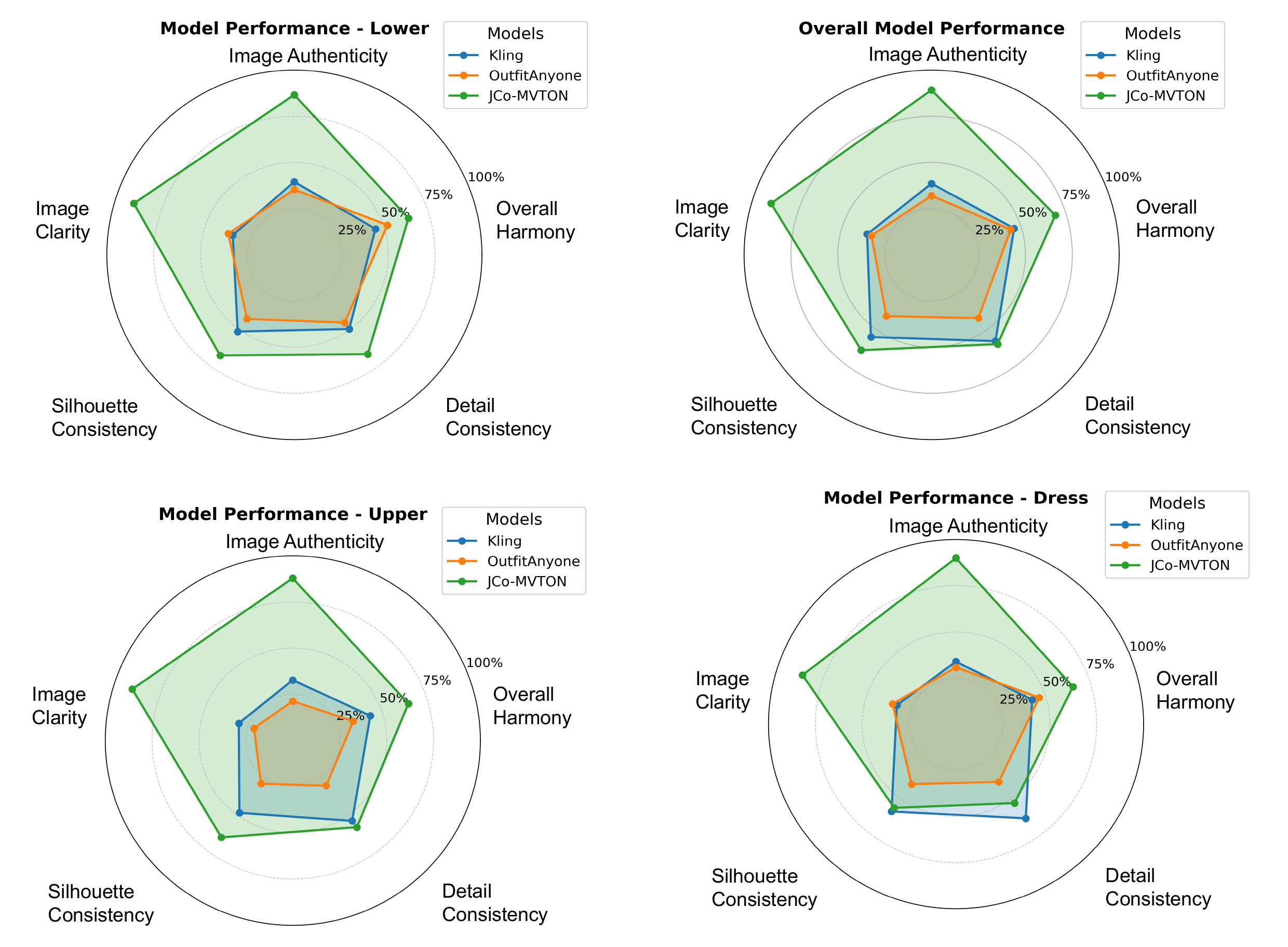}
    \caption{Radar‐chart comparison of three models (Kling, OutfitAnyone, and Ours JCo-MVTON) across four virtual try‐on scenarios:  
  (a) Overall: JCo-MVTON achieves the highest scores on all five metrics;  
  (b) Upper: JCo-MVTON leads on every metric;  
  (c) Lower: JCo-MVTON tops all metrics;  
  (d) Dress:   (c) Lower: JCo-MVTON tops all metrics;   attains the best performance on all metrics except a slight lag behind Kling in Detail Consistency.} % 添加图片标题
    \label{fig:my_example} % 添加标签，用于在文中引用
\end{figure}
\begin{itemize}
    \item \textbf{Image Authenticity}: Assesses how visually similar the generated image is to a real image, considering color, lighting, and background realism.
    \item \textbf{Image Clarity}: Evaluates the clarity of the generated image, checking for blurriness, noise, and other unwanted visual artifacts.
    \item \textbf{Silhouette Consistency}: Measures the consistency of the garment’s silhouette with the body shape, ensuring that the virtual clothing fits naturally.
    \item \textbf{Detail Consistency}: Assesses the accuracy of clothing details, such as fabric texture, wrinkles, and patterns, and whether they appear natural.
    \item \textbf{Overall Harmony}: Evaluates the overall visual harmony, considering how well the clothing and body integrate, and whether the outfit looks aesthetically pleasing.
\end{itemize}

We compared three models: \textbf{Kling}, \textbf{OutfitAnyone}, and \textbf{Ours JCo-MVTON}. These models represent different virtual try-on implementations, and we performed a human assessment for each model across various try-on scenarios. The specific evaluation scenarios are:

\begin{itemize}
    \item \textbf{Overall}: A holistic evaluation considering all metrics combined.
    \item \textbf{Upper}: Focused on the performance of upper-body clothing, such as shirts, blouses, and jackets.
    \item \textbf{Lower}: Focused on lower-body clothing, such as pants, skirts, and shorts.
    \item \textbf{Dress}: Focused on full-body clothing, especially dresses.
\end{itemize}

To facilitate the analysis and comparison of the results, we present radar charts for each scenario. The radar charts visually depict the scores of each model across the five evaluation metrics, providing a clear comparison of their performances in different scenarios.
\begin{figure}[h!] 
    \centering % 图片居中
    \includegraphics[width=0.9\textwidth]{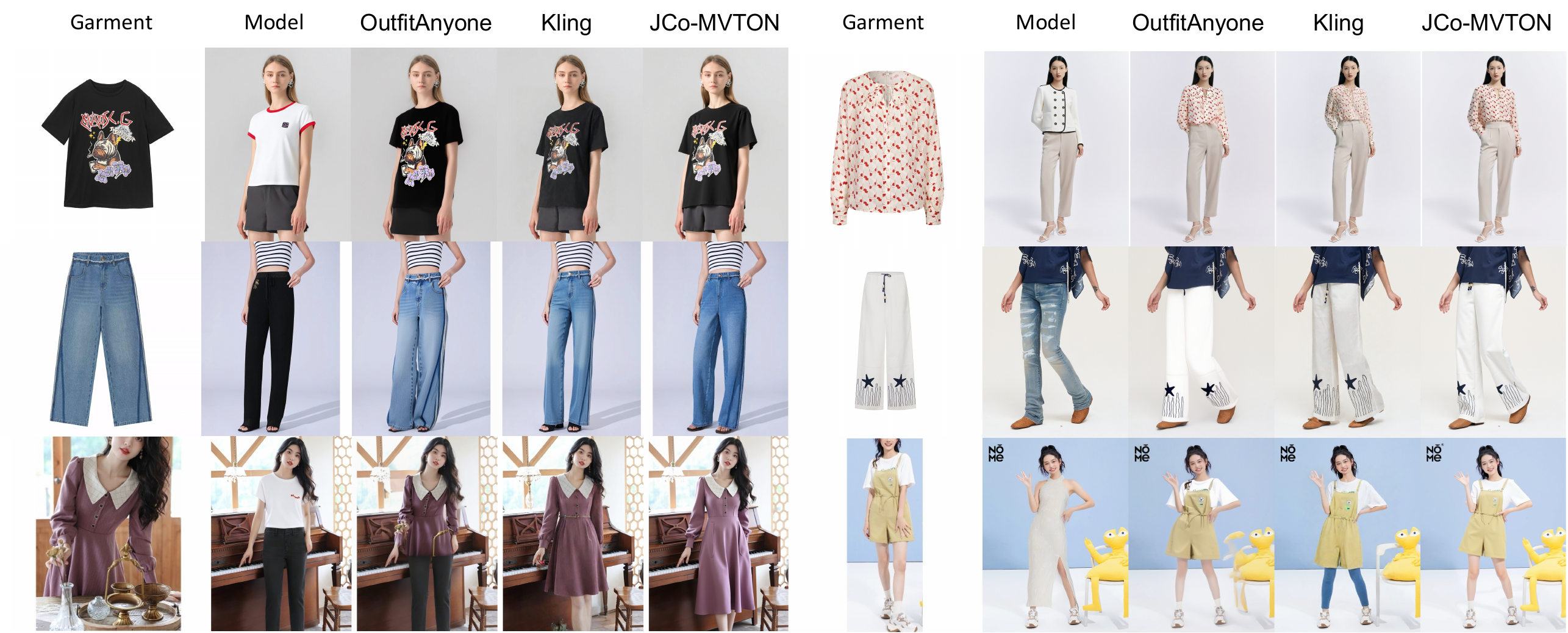}
    \caption{The manual evaluation results were derived from a sampled set of outcomes across upper-body, lower-body, and dress (full-body) tasks. The results, presented from left to right, reflect the performance of OutfitAnyone, Kling, and JCo-MVTON, respectively. According to our evaluation metrics, JCo-MVTON demonstrates consistently superior performance.} % 添加图片标题
    \label{fig:human_test} % 添加标签，用于在文中引用
\end{figure}

\subsection{Radar Chart Analysis}

\begin{itemize}
    \item \textbf{Overall}: Our model (\textit{JCo-MVTON}) outperforms Kling and OutfitAnyone, particularly in \textit{Image Clarity} and \textit{Image Authenticity}.
    \item \textbf{Upper}: In the upper-body clothing scenario, \textit{JCo-MVTON} slightly outperforms Kling in Detail Consistency, but achieving much higher scores in other metrics.
    \item \textbf{Lower}: In the lower-body clothing scenario, \textit{JCo-MVTON} excels again, much better on all five metric .
    \item \textbf{Dress}: For full-body clothing, \textit{JCo-MVTON} shows the strongest \textit{Overall Harmony} although it falls slightly short of Kling in \textit{Detail Consistency}.
\end{itemize}

\subsection{Discussion and Conclusion}

Based on the radar chart analysis, our model (\textit{Ours}) consistently outperforms the other models across all scenarios, particularly in terms of \textit{Image Authenticity}, \textit{Detail Consistency}, and \textit{Overall Harmony}. This indicates that \textit{Ours} is better at handling the fusion of clothing and the human body, producing more natural and harmonious virtual try-on results.

Although Kling and OutfitAnyone exhibit some advantages in specific scenarios, their performance in detail preservation and overall harmony remains limited. Therefore, based on these evaluation results, we conclude that \textit{Ours} demonstrates superior practicality and reliability in virtual try-on tasks.

%% file: sec/2_relatedwork.tex
\section{Related Works}
\subsection{Image-based Conditional Generation} 
Recent advances in image-conditioned generation have significantly improved both the fidelity and controllability of diffusion-based models.\citep{zhang2025easycontroladdingefficientflexible}proposed EasyControl, introducing a lightweight plug-and-play adapter for Diffusion Transformers (DiT)\citep{peebles2023scalablediffusionmodelstransformers} that injects image conditions via a parallel Low-Rank Adaptation (LoRA)\citep{hu2021loralowrankadaptationlarge} branch. By processing conditional inputs in isolation and employing separable convolutions with multi-scale feature aggregation, EasyControl achieves high-resolution outputs with flexible spatial and subject control, supporting zero-shot combination of controls at inference time. Shortly thereafter, Ominicontrol and Ominicontrol 2\citep{xie2024omnicontrolcontroljointtime,tan2025ominicontrol2efficientconditioningdiffusion}, which reuses the DiT’s own VAE encoder and multi-modal attention to embed image conditions with negligible extra parameters. OmniControl constructs a unified token sequence containing noisy image latents, text, and condition tokens, enabling the pretrained DiT to directly “wear” the control and handle both spatially-aligned inputs (e.g., edges, depth maps) and non-aligned subject images within a single model. Building on these, \citep{wang2025unicombineunifiedmulticonditionalcombination}introduced UniCombine, a DiT-based architecture for multi-conditional generation. UniCombine incorporates Condition-MMDiT Attention and multiple LoRA modules to jointly fuse text prompts, spatial maps, and reference images, offering both training-free and fine-tuned variants, and demonstrating state-of-the-art alignment across all input constraints on the new SubjectSpatial200K dataset.

% Example: place within your document preamble or body as fits your structure
\subsection{Mask-based and Mask-free Virtual Try-On}

In the past two years, image-based virtual try-on methods have bifurcated into mask-based and mask-free paradigms, each offering distinct advantages.
StableGarment\citep{wang2024stablegarmentgarmentcentricgenerationstable} pioneered the integration of a stability regularization term into the try-on pipeline, employing a two-stage framework that first predicts a coarse garment warped to the target shape and then refines details via a stability-aware generator to reduce artifacts along garment edges. Building on multi-view consistency, MV-VTON\citep{ wang2025mvvtonmultiviewvirtualtryon} incorporates multiple camera perspectives at training time: by enforcing cross-view feature alignment and view-consistent synthesis, it produces more accurate 3D-aware garment draping, particularly under challenging poses. LaDI-VTON\citep{morelli2023ladivtonlatentdiffusiontextualinversion} introduces a layered deformation initiative, decomposing garment transfer into a local alignment stage—where a dense correspondence field aligns fine-grained garment parts—and a global deformation network that adjusts the warped clothing to the body silhouette; this layered design significantly improves local detail preservation around collars and cuffs. DCI-VTON\citep{Gou_2023} further enhances local detail by proposing a Dual-Channel Integration module that fuses semantic segmentation and dense flow channels, enabling the network to exploit both structural layout and pixel-level correspondence cues; experiments demonstrate marked gains in rendering fidelity for complex patterns and textures. CatVTON\citep{chong2025catvtonconcatenationneedvirtual} proposes a category-aware transfer mechanism, in which garments are first classified into semantic categories (e.g., tops, bottoms, outerwear) and then passed through category-specific deformation subnets, resulting in improved adaptability across diverse garment types. More recently, OOTDiffusion\citep{xu2024ootdiffusionoutfittingfusionbased} applies diffusion-based generative modeling to the try-on task, formulating garment transfer as a conditional denoising process that iteratively refines a noisy composite of the person and target clothing; this approach excels at generating high-frequency textures and natural shading transitions but can incur greater computational cost due to the multi-step diffusion schedule. BooW-VTON\citep{zhang2024boowvtonboostinginthewildvirtual} is a mask-free virtual try-on method that uses augmented data instead of segmentation masks. It combines a U-Net with pre-trained models to encode clothing features and uses cross-attention to apply them. A special loss helps focus on the clothing area, reducing errors. This improves alignment and texture, even with difficult poses and lighting. MF-VITON\citep{wan2025mfvitonhighfidelitymaskfreevirtual} removes the need for user-provided masks with a two-stage, mask-free approach. First, it uses a mask-based model to create a large, diverse dataset with different backgrounds. Then, it fine-tunes the model to work without masks. This allows garment transfer using just one person image and a target garment, while keeping texture and shape details and achieving top performance. Nevertheless, while current approaches remain limited by either architectural complexity or insufficient control over virtual try-on synthesis, they collectively inform pathways toward more efficient and streamlined implementations.

%% file: sec/3_method.tex
\section{Jointly Conditional MM-DiT Model for Mask-Free Vitual Try-on  }
\noindent
In this section, we provide an overview of the Jointly Controllable MMDiT, which achieves high-quality, stable, mask-free virtual try-on by jointly injecting multiple image features (e.g., reference image and garment image) into the self-attention mechanism.

\subsection{Preliminary} \label{sec:predliminary}
\paragraph{Multimodal Diffusion Transformers} Multimodal Diffusion Transformers (MM-DiT)\citep{esser2024scalingrectifiedflowtransformers} is a text-driven image generation architecture, commonly used in text-to-image models such as FLUX\citep{flux2024} and SD3\citep{esser2024scalingrectifiedflowtransformers}. It can effectively integrate textual and visual features into the attention architecture, making it especially well‑suited for multi‑image–controlled virtual try-on (VTON) tasks. The model leverages diffusion processes to generate visual content while employing transformer-based structures for cross-modal information fusion, enabling effective alignment between clothing and the human body, thereby enhancing the quality and stability of virtual try-on results. In MM-DiT, the diffusion model generates the target image based on a conditional diffusion process. Let \( x_0 \) represent the initial state of the generated image (i.e., starting from noise), and the true image is gradually restored through the diffusion process. The training objective of the model is to minimize the following loss function:
\begin{equation}
L_{\text{diff}} = \mathbb{E}_{q(x_t \mid x_{t-1})} \left[ \| x_t - f_{\theta}(x_{t-1}, c) \|^2 \right]
\end{equation}
where \( x_t \) represents the image representation at step \( t \) in the diffusion process, \( f_{\theta} \) is the transformer-based generative network, and \( c \) is the conditional information (such as body features, clothing features, etc.). Through this process, the model can generate high-quality virtual try-on images based on different conditional information.

\paragraph{Rectified Flow for Diffusion-Based Generation} Rectified Flow (RF)~\citep{liu2022flowstraightfastlearning} is a reformulation of the traditional diffusion framework, designed to simplify the learning dynamics of generative models. Instead of modeling a complex score function or simulating stochastic processes, RF directly learns a \textit{deterministic vector field} that transports noisy samples back to data points in a continuous and efficient manner. This deterministic perspective enables the generation process to be framed as solving an ordinary differential equation (ODE), thus reducing the complexity of sampling.

In standard score-based diffusion models, a sample $x_t$ is generated by gradually adding noise to the data $x_0$, and then denoised via the learned score function $\nabla_{x_t} \log p_t(x_t)$. In contrast, Rectified Flow directly models the velocity field $\mathbf{v}_\theta(x_t, t)$ that maps $x_t$ toward $x_0$ for $t \in [0, 1]$, where $t$ is a continuous time variable.

The training objective of Rectified Flow is to minimize the \textit{velocity matching loss}:

\begin{equation}
\mathcal{L}_{\text{RF}} = \mathbb{E}_{x_0 \sim p_{\text{data}}, \epsilon \sim \mathcal{N}(0, I), t \sim \mathcal{U}(0, 1)} \left[ \left\| \mathbf{v}_\theta(x_t, t) - \frac{x_0 - x_t}{1 - t} \right\|^2 \right]
\end{equation}

where $x_t$ is constructed as a linear interpolation between data and noise:

\begin{equation}
x_t = (1 - t) x_0 + t \epsilon
\end{equation}

This formulation allows $x_t$ to smoothly interpolate between clean data $x_0$ and Gaussian noise $\epsilon$, while the model learns to output a velocity vector that correctly points from $x_t$ to $x_0$, scaled by the temporal factor $(1 - t)^{-1}$.

At inference time, samples are generated by solving the following ODE:

\begin{equation}
\frac{dx}{dt} = \mathbf{v}_\theta(x, t)
\end{equation}

Compared to conventional stochastic diffusion processes, Rectified Flow enables a more efficient and stable sampling procedure. This property makes it especially suitable for high-fidelity generation tasks such as virtual try-on (VTON), where stability, sample quality, and alignment between modalities are crucial.

\subsection{Data Preparation} \label{sec:training}
\noindent

The objective of our pipeline is to construct a large-scale, high–fidelity
mask‑free virtual try‑on dataset composed of triplets
$\{G,P,R\}$,
where $G$ is a garment image, $P$ is a source person image,
and $R$ is the reference (post‑try‑on) person image.
We adopt a two–stage strategy that progressively improves both data quality
and domain diversity.

%-----------------------------------------------------------
\subsubsection{Stage I – Data Collection}

\paragraph{I.a Data Collection.}
We start from two public benchmarks---VITON~\citep{han2018viton} and
DressCode~\citep{morelli2022dresscodehighresolutionmulticategory}---and crawl additional Internet images.
The raw pool contains isolated person images $\{P\}$ as well as
paired garment–person examples $\{G,P\}$.
Isolated images significantly enrich pose, identity, and lighting
variation, but an explicit garment representation is missing.
\begin{figure*}[t]
    \centering
    \includegraphics[width=\textwidth]{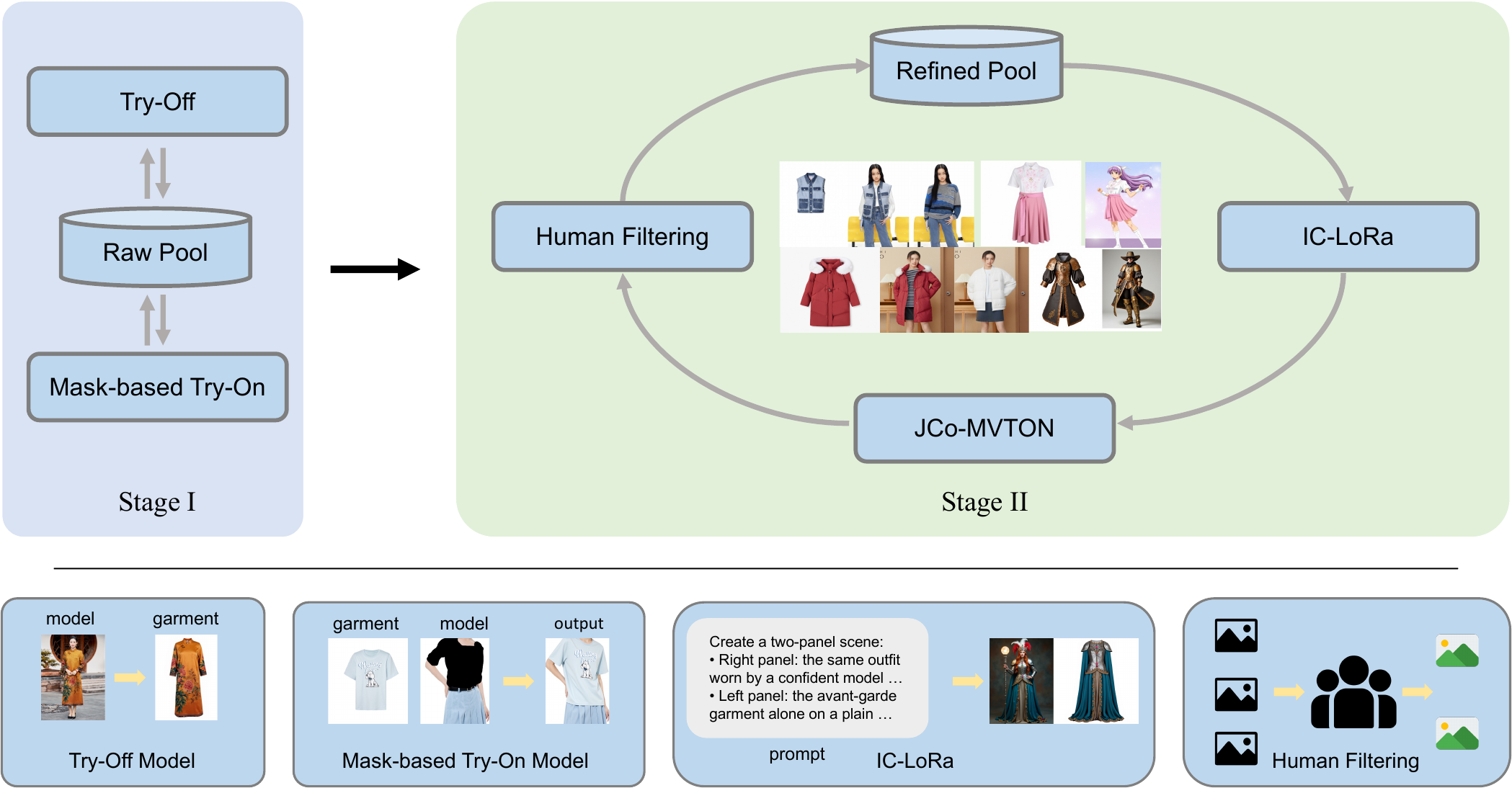}
% \vskip -0.05in
\caption{Overview of the two‑stage data pipeline used to build our mask‑free virtual try‑on dataset. Stage I bootstraps a raw pool by alternately running a Try‑Off model, which generates garment images, and a mask‑based try‑on model, which produces initial reference images. Stage II iteratively refines and enlarges the dataset: human‑in‑the‑loop filtering cleans the seed pool, JCo‑MVTON regenerates sharper triplets, and ICLoRA injects new styles to widen the domain. The cycle repeats until the refined pool reaches the desired quality and diversity.}
\label{fig:data}
% \vspace{-5mm}
\end{figure*} 

\paragraph{I.b Try‑Off garment recovery}
To infer $G$ from a sole person image $P$,
we train a Try‑Off diffusion model that mirrors the architecture of
our JCo-MVTON but swaps the generation target to the garment image.
Training pairs are obtained from the paired $\{G, P\}$ subset,
where the garment foreground is extracted with BiRefNet\citep{zheng2024bilateral}.
Because Try‑Off task is mask‑free and leverages the strong generative
prior of FLUX, it generalizes well to challenging poses, occlusions, and
lighting conditions.

\paragraph{I.c Mask‑based Try‑On.}
The recovered pairs $\{G,P\}$ pairs are fed to a FLUX‑Fill model
that performs mask‑guided try‑on. For each person image $P$ we randomly sample a garment $G'$ and obtain corresponding $R'$.
This produces the first coarse triplet set that bootstraps our
mask‑free model.

%-----------------------------------------------------------
\subsubsection{Stage~II: Quality and Domain Refinement}

\paragraph{II.a Human–in–the–loop Filtering.}
Triplets are scored along three axes:
(i) \textbf{Garment Consistency} ($G$ vs.\ $P$ in garment region),
(ii) \textbf{Person Consistency} ($P$ vs.\ $R$ in pose, identity, body shape),
and (iii) \textbf{Photorealism} of $R$.
We employ professional annotators and an interactive GUI to retain only
high‑quality samples.

\paragraph{II.b ICLoRA‑based Domain Expansion.}
To broaden style coverage, we employ In‑Context LoRA (ICLoRA) \citep{huang2024context}
to fine‑tune the frozen FLUX backbone with low‑rank adapters
on the filtered $\{G,P\}$ pairs
Prompt engineering with an LLM then drives the model to synthesize
novel style‑aware pairs, ranging from anime style to
cyber‑punk outfits, largely enriching the domain distribution.

\paragraph{II.c Iterative Mask‑free Bootstrapping.}
The curated triplets train our first–round JCo-MVTON model.
Owing to its flexible conditioning (no segmentation mask),
JCo-MVTON can now regenerate higher‑fidelity $R$ images,
especially for failure cases discovered in Stage~I.c.
The regenerated triplets are re‑filtered via II.a,
yielding an expanded corpus with both better quality and
wider coverage.
We iterate II.a–II.c until performance saturates
(three rounds in practice, resulting in $\sim120$K triplets).
\subsection{Multi-Conditional MMDiT Attention on Virtual Try-on }
\noindent

\begin{figure*}[t]
    \centering
    \includegraphics[width=\textwidth]{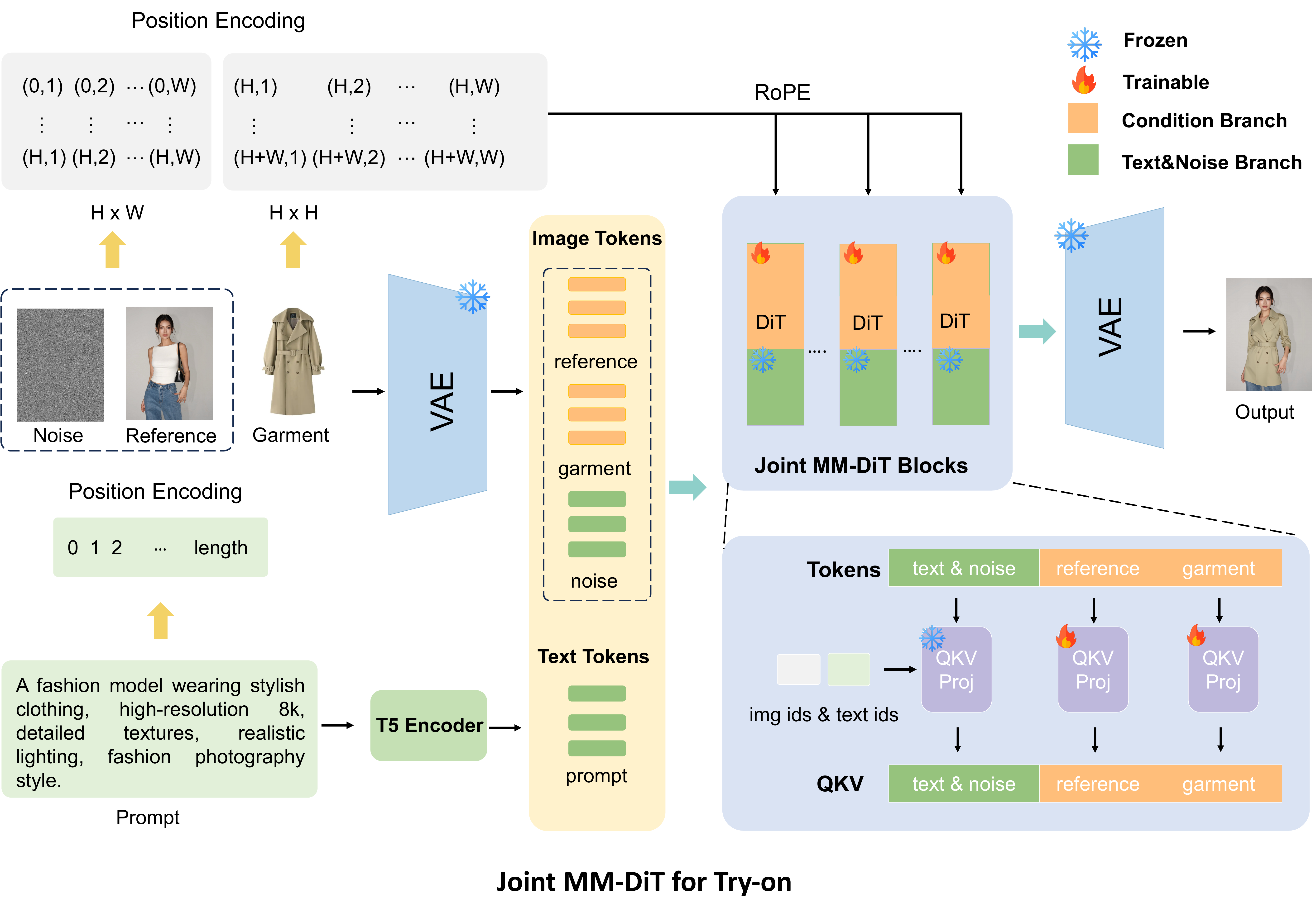}
% \vskip -0.05in
\caption{The overall framework of JCo-MVTON. Given a fixed prompt, reference image, and garment image as inputs, the Joint MM-DiT architecture fuses multi-conditional features to synthesize the try-on image. Within each Joint MM-DiT block, noise and conditional features are processed through three parallel branches and fused via self-attention.}
\label{fig:main}
% \vspace{-5mm}
\end{figure*}

As shown in Figure \ref{fig:main}, JCo-MVTON builds upon the FLUX\citep{flux2024} architecture. FLUX injects noisy image embeddings alongside textual prompts into its MM-DIT self-attention layers, leveraging cross-modal interactions to iteratively guide and refine the synthesis process, thereby generating images that faithfully reflect the input text. Inspired by this approach, we propose augmenting the conditioning signal by directly injecting VAE-encoded image latent representations into the self-attention layers. By injecting diverse information as multi-conditional inputs into the self-attention layers, we enable seamless incorporation of various control signals, facilitating high-fidelity, controllable image generation.

\paragraph{Conditional Self-Attention for Try-on.}
In the mask-free virtual try-on framework, we first encode the text prompt, random noise, reference image, and garment image using a text encoder and a variational auto-encoder (VAE), respectively, to obtain their corresponding intermediate feature maps. These feature maps are then projected into a shared embedding space through an embedding network, yielding embedding vectors 
\( T \), \( X \), \( C_1 \), and \( C_2 \). Subsequently, the embeddings are concatenated along the channel dimension to form a unified embedding sequence: \begin{equation} S = [T; X; C_1; C_2], \end{equation} which serves as the conditional input for the subsequent self-attention mechanism to enable coherent garment synthesis and pose transfer.

In order to preserve the fundamental capabilities of FLUX, the newly injected conditional inputs, such as the reference image, garment image, are processed through the same VAE and embedding layers as the original data. The only modification introduced is the use of two separate query, key, and value (QKV) projection branches during the self-attention operation. These branches are initialized with the weights from the original text \& noise projection branch and are updated during training.

During self-attention computation, the sequences of prompt and noise features are first projected through the primary branch to produce \((Q_{1},K_{1},V_{1})\). Similarly, the reference and garment feature sequences are projected via the conditional branch to produce \((Q_{2},K_{2},V_{2})\) and \((Q_{3},K_{3},V_{3})\), respectively.  Formally, this projection step can be written as
\begin{equation}
(Q_{i},\,K_{i},\,V_{i})
=
\bigl(W_{Q}Z_{i},\,W_{K}Z_{i},\,W_{V}Z_{i}\bigr),
\quad i\in\{t\&n,\;c_{1},\;c_{2}\}.
\end{equation}

\paragraph{Masked Attention Mechanism for Try-on.}

Masked self-attention enforces a causality constraint within each individual QKV head before concatenation, ensuring that the multi-head attention mechanism, as a whole, does not utilize information from future tokens in the sequence. In the context of the try-on task, particularly when dealing with multiple conditional inputs, there is often no direct relationship between the features of the reference image and the garment image. When multi-head attention is applied without considering this, it can lead to suboptimal performance by introducing unwanted interactions between the two input sources. To address this, it is essential to mask the interactions between features from different conditions, thereby preventing them from attending to each other during the attention process. This masking mechanism allows for independent attention among text, noise, and individual conditions, while ensuring that the conditions themselves do not cross-attend. The resulting model not only enhances the generative performance for the try-on task but also helps reduce computational time complexity, as it limits unnecessary interactions during attention computation.

Let \(M\) be the binary mask that controls the allowable attention between conditional branches. We define
\begin{equation}
Q = \bigl[\,Q_{t\&n},\;Q_{c1},\;Q_{c2}\bigr], 
\quad
K = \bigl[\,K_{t\&n},\;K_{c1},\;K_{c2}\bigr], 
\quad
V = \bigl[\,V_{t\&n},\;V_{c1},\;V_{c2}\bigr].
\end{equation}
where \(Q_{t\&n}, K_{t\&n}, V_{t\&n}\) correspond to the \emph{text \& noise} branch, \(Q_{c1}, K_{c1}, V_{c1}\) correspond to the \emph{reference-image} branch, and \(Q_{c2}, K_{c2}, V_{c2}\) correspond to the \emph{garment-image} branch.

For each position $i$, we compute raw attention scores against each position $j$:
\begin{equation}
\text{score}_{ij} = \frac{Q_i \cdot K_j^T}{\sqrt{d_k}}.
\end{equation}
We then apply the mask \(M \in \mathbb{R}^{n\times n}\) defined by
\begin{equation}
M_{ij} = 
\begin{cases}
-\infty, & (i\in c_1,\,j\in c_2)\lor(i\in c_2,\,j\in c_1),\\
0,        & \text{otherwise}.
\end{cases}
\end{equation}

Adding \(M\) to the score matrix \(S\) effectively sets any \(S_{ij}\) with \((i \in c_1,\, j \in c_2) \lor (i \in c_2,\, j \in c_1)\) to \(-\infty\), so that after the softmax they receive zero weight.

\begin{equation}
\mathrm{softmax}(S + M)_{ij}
= \frac{\exp(S_{ij} + M_{ij})}{\sum_{k=1}^n \exp(S_{ik} + M_{ik})}.
\end{equation}

In multi-condition try-on tasks, a mutually exclusive attention mask is introduced to constrain self-attention so that each condition branch attends only to itself or to explicitly permitted branches, thereby preventing interference and noise propagation. This mechanism not only preserves each branch’s internal feature modeling capacity—enhancing the accuracy and robustness of the generated results—but also reduces computational and memory overhead by eliminating invalid attention computations. By flexibly defining the mask matrix, the approach can be extended to additional condition branches or more complex combinations, striking a balance between branch-specific modeling and necessary cross-condition fusion.

\subsection{Controllable Positional Encoding for Try-on } \label{sec:dynamic_token}
\noindent
Transformers fundamentally rely on positional encodings to distinguish among tokens or patches, yet conventional fixed or learned absolute embeddings often fail to generalize across varying sequence lengths or spatial resolutions. In contrast, the FLUX framework adopts rotary positional embeddings (RoPE) \citep{su2023roformerenhancedtransformerrotary} to introduce a natural relative‐position bias: by applying a rotation in the query/key vector space, RoPE inherently encodes inter‐token relationships and thus seamlessly supports inputs of arbitrary length or resolution.

\paragraph{Image Positional Encoding}
In the first step of our try-on model, the input image of spatial dimensions \(H \times W\) is divided into a grid of non‐overlapping patches, each of size \(P \times P\), yielding a total of
\[
N \;=\; \frac{H}{P} \times \frac{W}{P}
\]
patches. Each patch is then flattened into a vector of length \(P^2 \cdot C\), where \(C\) is the number of input channels, and subsequently projected through a learnable linear layer to produce a fixed‐dimensional embedding of size \(d\).

Formally, if we denote the set of flattened patches by
\[
\{\mathbf{p}_i\}_{i=1}^N,
\]
the patch embedding operation can be written as
\[
\mathbf{x}_i \;=\; \mathbf{W}_{\mathrm{emb}}\,\mathbf{p}_i + \mathbf{b}_{\mathrm{emb}}
\;\;\text{for}\;\; i = 1,\dots,N,
\]
where
\[
\mathbf{W}_{\mathrm{emb}}
\;\in\;\mathbb{R}^{d \times (P^2 C)},
\quad
\mathbf{b}_{\mathrm{emb}}
\;\in\;\mathbb{R}^d
\]
are parameters learned during training. This yields a sequence of patch embeddings
\(\{\mathbf{x}_i\}_{i=1}^N\), each of dimension \(d\), which serves as the input token sequence for the subsequent Transformer layers.

\paragraph{Try-on based Jonit Positional Encoding.}

In mask-free virtual try-on tasks, the choice of positional encoding is critical. We compare two schemes:
\begin{itemize}
  \item \textbf{Scheme I:} Adopt a unified positional space for the prompt, noise, reference image, and garment image.
  \item \textbf{Scheme II:} Recognizing that in try-on scenarios the noise and reference image share identical spatial dimensions and backgrounds, assign them a common positional space. Inspired by CatVTON\citep{chong2025catvtonconcatenationneedvirtual}, we then incorporate the garment image into this space by concatenation, thereby preventing interference with the background of the generated output.
\end{itemize}
In our scheme, we assume the noise and reference images have spatial dimensions \((H, W)\), while the garment image has dimensions \((H, H)\). Their positional encodings in sequence are:

\[
(0,0),\,(0,1),\dots,(H,W)
\]

for the noise and reference images, and

\[
(0,W),\,(0,W+1),\dots,(H,W+H)
\]

for the garment image.

Our experiments demonstrate that this concatenation-based positional encoding strategy is better suited to virtual try-on tasks, yielding stronger background preservation and higher-quality garment swapping results.

\subsection{Training Strategy} \label{sec:training}
\noindent

\paragraph{Coarse-to-fine training.}
For complex tasks such as virtual try-on, it is difficult for the FLUX\citep{flux2024} base model to produce a high-resolution result in a single training pass. Inspired by the approaches of \citep{karras2018progressivegrowinggansimproved,ho2021cascadeddiffusionmodelshigh} et al, we therefore first train a low-resolution (512×512) try-on model, and subsequently fine-tune it to obtain a more stable high-resolution (1024×1024) variant. To minimize disruption of the FLUX model’s original capabilities, we freeze all of its pre-existing parameters and train only an auxiliary QKV branch. Two strategies are explored for integrating this branch: one based on LoRA and the other via full-parameter fine-tuning. Our comparative experiments demonstrate that full-parameter training yields superior performance.

\paragraph{Fine-Tuning with Augmented Data.}
To address the try-on model's inadequate performance in certain task-specific scenarios, we conducted targeted post-training fine-tuning on the pretrained model, thereby ensuring its strong generalization across multiple contexts. Using the IC-LoRa technique, we generated augmented data that encompassed rare scenarios and uncommon garment types. Experimental results show that training with only 500–1,000 sets of such targeted augmented samples is sufficient to achieve strikingly improved performance.

%% file: sec/4_experiment.tex
\section{Experiments}
\paragraph{Implementation details.}
The model was initialized from the original FLUX.1-dev\citep{flux2024} weights with two additional QKV\citep{vaswani2023attentionneed} projection branches and trained on eight NVIDIA H20 GPUs using the Prodigy\citep{mishchenko2024prodigyexpeditiouslyadaptiveparameterfree} optimizer at a learning rate of 1 in two stages—first at 512×384 resolution with a batch size of 16, then at 1024×768 resolution with a batch size of 4.

\paragraph{Datasets.}   
We have cumulatively collected tens of millions of e-commerce data related to clothing, and then performed basic data preprocessing such as cleaning, deduplication, and filtering. Finally, the training dataset comprises a total of 141,734 high-quality triplets (each consisting of a model, a reference image and a garment), spanning three scenarios: upper-body, lower-body and full-body. Specifically, it contains 69,261 upper-body triplets, 33,838 lower-body triplets and 38,635 dress triplets. All samples were initially generated by IC-LoRA\citep{huang2024context} and mask-based try-on networks, and subsequently subjected to manual filtering to ensure quality.                                        

\paragraph{Metrics.}
In paired try-on scenarios where test sets include ground-truth images, we assess the similarity between generated and real images using four standard metrics—Structural Similarity Index (SSIM) \citep{1284395}, Learned Perceptual Image Patch Similarity (LPIPS) \citep{zhang2018unreasonableeffectivenessdeepfeatures}, Frechet Inception Distance (FID)\citep{Seitzer2020FID} and Kernel Inception Distance (KID)\citep{bińkowski2021demystifyingmmdgans}. For unpaired settings, we instead compare the distributions of generated and real samples using FID and KID.

\begin{table}[h!]
\centering
\caption{Quantitative comparison with other methods on VITON-HD\citep{choi2021vitonhdhighresolutionvirtualtryon}  dataset. The best results are demonstrated in \textbf{bold}.}
\label{tab:comparison_viton_hd}
\begin{tabular}{l|cccc|cc}
\toprule
\textbf{Methods} & \multicolumn{4}{c|}{\textbf{Paired}} & \multicolumn{2}{c}{\textbf{Unpaired}} \\
\cmidrule(lr){2-5} \cmidrule(lr){6-7}
& SSIM $\uparrow$ & FID $\downarrow$ & KID $\downarrow$ & LPIPS $\downarrow$ & FID $\downarrow$ & KID $\downarrow$ \\
\midrule
StableGarment \citep{wang2024stablegarmentgarmentcentricgenerationstable} & 0.8029 & 15.567 & 8.519 & 0.1042 & 17.115 & 8.851 \\
MV-VTON \citep{wang2025mvvtonmultiviewvirtualtryon} & 0.8083 & 15.442 & 7.501 & 0.1171 & 17.900 & 3.861 \\
LaDI-VTON \citep{morelli2023ladivtonlatentdiffusiontextualinversion} & 0.8603 & 11.386 & 7.248 & 0.0733 & 14.648 & 8.754 \\
DCI-VTON \citep{Gou_2023} & 0.8620 & 9.408 & 4.547 & 0.0606 & 12.531 & 5.251 \\
OOTDiffusion \citep{xu2024ootdiffusionoutfittingfusionbased} & 0.8187 & 9.305 & 4.086 & 0.0876 & 12.408 & 4.689 \\
GP-VTON \citep{xie2023gpvtongeneralpurposevirtual} & \textbf{0.8701} & 8.726 & 3.944 &\textbf{0.0585} & 11.844 & 4.310 \\
JCo-MVTON (Ours) &0.8601 &\textbf{8.103} &\textbf{2.003} &0.0891 &\textbf{9.561} &\textbf{2.700} \\
\end{tabular}
\end{table}

\subsection{Comparison with State-of-the-Art Try-on Models}
To quantitatively evaluate the performance of our try-on model on garment-exchange tasks, we conducted experiments on two datasets—VITON-HD\citep{choi2021vitonhdhighresolutionvirtualtryon} and DressCode\citep{morelli2022dresscodehighresolutionmulticategory}. For each dataset, we trained a dedicated version of the try-on model and carried out independent tests at a resolution of 1024 × 768. The VITON-HD dataset comprises only upper-body garments, whereas the DressCode dataset includes upper-body, lower-body, and dress clothing. Moreover, both datasets provide paired and unpaired configurations for evaluation.

As shown in Table \ref{tab:comparison_viton_hd}, we compare the performance of JCo-MVTON on the VITON-HD dataset. Owing to its mask-free design, our model substantially outperforms all competing approaches under the unpaired evaluation protocol and achieves state-of-the-art or near–state-of-the-art results under the paired evaluation protocol, particularly surpassing all baselines on both FID and KID metrics. Table \ref{tab:dresscode_comparison} reports results on the DressCode dataset, which comprises three distinct garment categories—upper, lower, and dress. Again, JCo-MVTON consistently exceeds the performance of all compared methods, delivering outstanding results across every task category.

\begin{table}[ht]
\centering
\caption{Quantitative comparison on the DressCode \citep{morelli2022dresscodehighresolutionmulticategory}dataset for the upper-body, lower-body, and dress tasks. The best results are shown in bold.}
\label{tab:dresscode_comparison}
\begin{tabular}{@{}l cccccc@{}}
\toprule
\textbf{Model}    & \textbf{LPIPS $\downarrow$} & \textbf{SSIM $\uparrow$} & \textbf{FID$_p$ $\downarrow$} & \textbf{KID$_p$ $\downarrow$} & \textbf{FID$_u$ $\downarrow$} & \textbf{KID$_u$ $\downarrow$} \\ \midrule
\multicolumn{7}{l}{\textbf{Upper}} \\
LaDI-VITON        & 0.1091          & 0.9044          & 18.14           & 3.703           & 16.43           & 4.829           \\
OOTD              & 0.0855          & 0.8997          & 16.20           & 5.862           & 13.20           & \textbf{1.860}           \\
IDM-VTON          & 0.0761          & 0.9125          & 16.25           & 7.352           & 13.60           & 2.952           \\
\textbf{Ours}     &  \textbf{0.0695}         & \textbf{0.9123}          & \textbf{10.92}           & \textbf{3.022}           &\textbf{11.53}          & 2.574                 \\ \midrule
\multicolumn{7}{l}{\textbf{Lower}} \\
LaDI-VITON        & 0.1314          & 0.8855          & 14.98           & 3.920           & 13.95           & \textbf{2.564}           \\
OOTD              & 0.1168          & 0.8706          & 15.56           & 3.797           & 21.50           & 8.248           \\
IDM-VTON          & 0.1103          & 0.8869          & 18.20           & 8.123           & 15.97           & 5.386           \\
\textbf{Ours}     &\textbf{0.0721}           &\textbf{0.8913}           &\textbf{11.08}            &\textbf{2.569}            &\textbf{13.72}            & 3.83             \\ \midrule
\multicolumn{7}{l}{\textbf{Dress}} \\
LaDI-VITON        & 0.1753          & 0.8424          & 24.00           & 13.70           & 16.86           & 5.005           \\
OOTD              & 0.1490          & 0.8440          & 25.75           & 15.29           & 20.95           & 8.149           \\
IDM-VTON          & 0.1381          & 0.8627          & 25.96           & 17.67           & 12.82           & 9.739           \\
\textbf{Ours}     & \textbf{0.0732}          &\textbf{0.9032}           &\textbf{11.82}            &\textbf{2.942}            &\textbf{12.54}           &\textbf{3.576}            \\ \bottomrule
\end{tabular}
\end{table}

\begin{table}[t]
  \centering
  \caption{Detailed ablation study on the VITON-HD dataset: comparison of three configurations—without positional encoding, without updating all parameters, and our full model —evaluated on paired and unpaired  settings.}
  \label{tab:ablation_experiments_short}
  \begin{tabular}{lcccccc}
    \toprule
    \multirow{2}{*}{\textbf{Method}} 
      & \multicolumn{4}{c}{\textbf{Paired}} 
      & \multicolumn{2}{c}{\textbf{Unpaired}} \\
    \cmidrule(lr){2-5} \cmidrule(lr){6-7}
      & SSIM $\uparrow$ & FID $\downarrow$ & KID $\downarrow$ & LPIPS $\downarrow$ 
      & FID $\downarrow$ & KID $\downarrow$ \\
    \midrule
    w/o Pos Encoding
      & 0.85 &  9.04 & 3.18 & 0.09 
      & 10.20 & 2.77 \\
    w/o Full-Params
      & 0.84 & 10.54 & 4.08 & 0.06 
      & 11.49 & 4.76 \\
    \midrule
    Ours
      & \textbf{0.86} & \textbf{8.10} & \textbf{2.00} & \textbf{0.09}
      & \textbf{9.56} & \textbf{2.70} \\
    \bottomrule
  \end{tabular}
\end{table}

\subsection{Ablation Study}
\paragraph{Condition LoRA Branch}
To efficiently and conveniently incorporate conditional variables (i.e. reference and garment images), we propose extending MMDiT by branching its Q, K, and V projections. In our first scheme, we duplicate the original QKV projection branch, freeze all backbone weights, and train only the newly added branch. In the alternative scheme, we further augment the duplicated branch with a LoRA\citep{hu2021loralowrankadaptationlarge} adapter and optimize exclusively its low-rank parameters—leaving both the backbone and duplicated projections intact. By restricting updates to a small subset of parameters, the LoRA-based approach substantially reduces training complexity, accelerates convergence, and better preserves the integrity of the pretrained model.

\paragraph{Jonit Positional Encoding.}
The FLUX model employs positional encodings to represent the relative spatial locations of pixels. Notably, while the pixels of the target and reference images are spatially aligned with respect to the background, the garment exhibits little inherent spatial correspondence. To reinforce background consistency in the try-on task, we propose to synchronize the positional encodings of the noise input with those of the reference image, whereas the garment image’s positional encoding is constructed via horizontal concatenation. Experimental results demonstrate that this joint positional encoding scheme significantly enhances both the synthesis quality and background preservation capabilities of the model in the try-on task.
In Table \ref{tab:ablation_experiments_short} we can see that once we replace the original sinusoidal positional encoding with one that is learned to exactly match the characteristics of our generated outputs.Intuitively, because the reference model’s positions and the generated clothing‐selection positions now share the same embedding space, our simple "concatenate‐then‐decode" fusion no longer suffers from misaligned spatial priors; this alignment effectively suppresses the cross‐interference that previously blurred fine details around sleeve hems and neckline contours.
\begin{figure}[h!] % [h!] 选项建议LaTeX将图片放在代码所在位置 (here)
    \centering % 图片居中
    \includegraphics[width=0.9\textwidth]{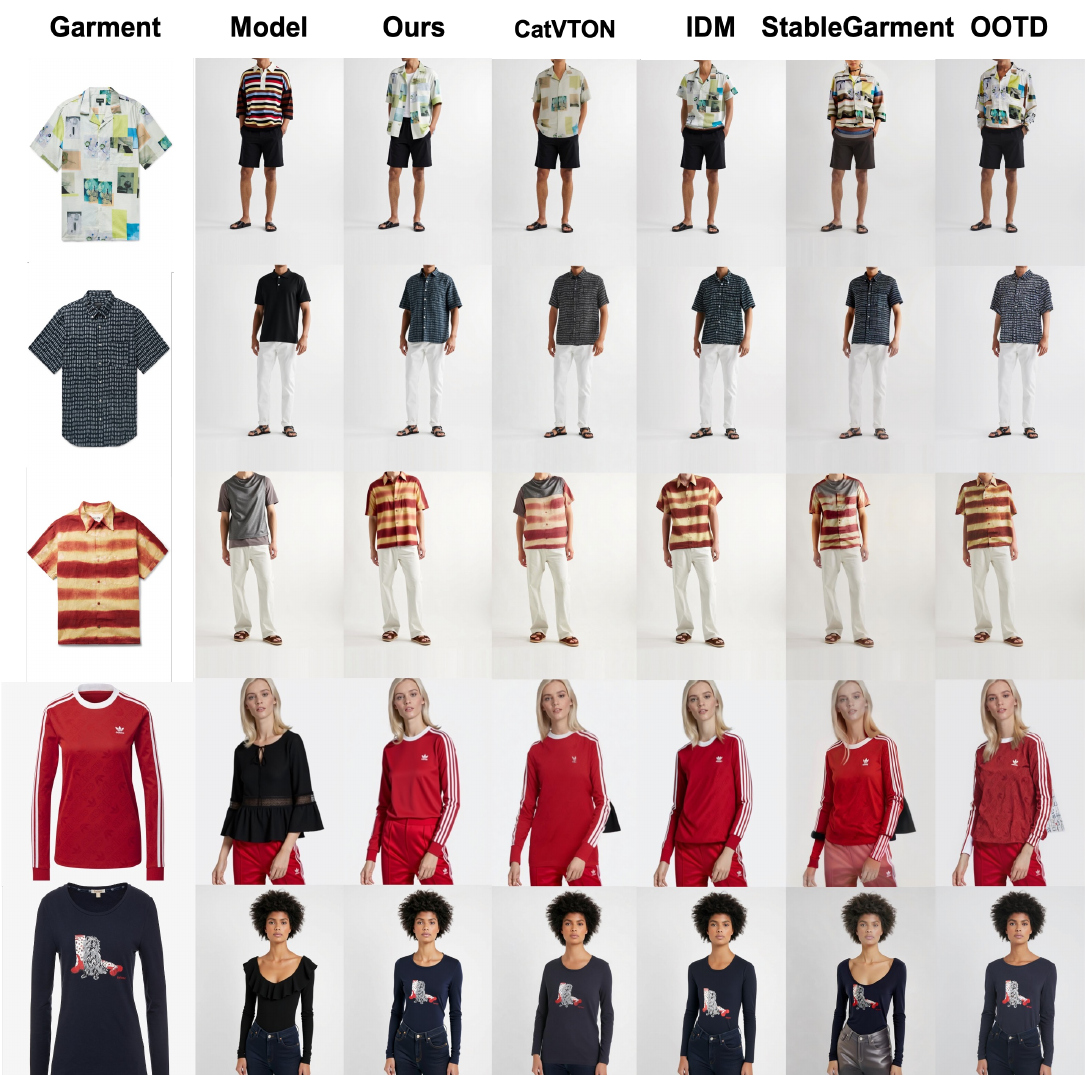}
    \caption{The figure presents high-resolution results of our try-on model across various scene-specific tasks, demonstrating its strong generalization capability and the consistently high quality of its outputs.} % 添加图片标题
    \label{fig:5} % 添加标签，用于在文中引用
\end{figure}
\begin{figure}[h!] % [h!] 选项建议LaTeX将图片放在代码所在位置 (here)
    \centering % 图片居中
    \includegraphics[width=0.9\textwidth]{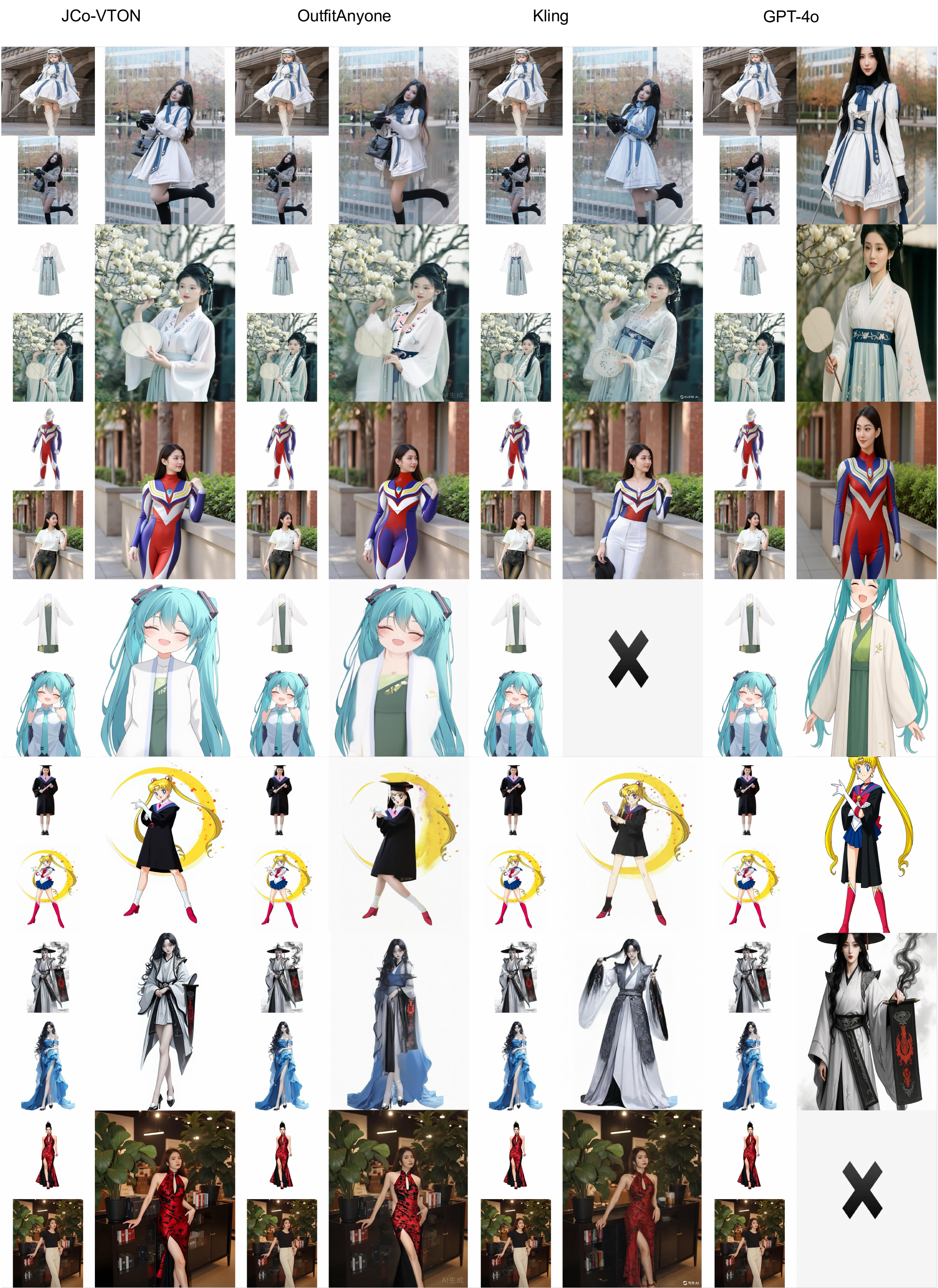}
    \caption{Comparison of our model and other commercial systems on wild-image scenarios. From left to right: our result, OutfitAnyone, Kling, and GPT-4o. The samples encompass real-person, anime, and classic styles. Across all style categories, our model delivers superior performance.} % 添加图片标题
    \label{fig:6} % 添加标签，用于在文中引用
\end{figure}

By contrast, the version that routes garment‐position signals through a LoRA branch underperforms dramatically. We attribute this deficit to LoRA’s inherently low‐rank adaptation, which—while parameter‐efficient—struggles to encode precise, per‐pixel spatial offsets. In practice, the branch’s limited capacity to generate sharply localized shifts results in "floating" artifacts around the boundary of the overlaid garment, suggesting that a more expressive injection mechanism (e.g., full dense fusion or higher‐rank adapters) would be necessary to regain fine‐grained control in the try-on pipeline.

\subsection{Visualization}
To more intuitively and vividly demonstrate our virtual try-on model’s performance across diverse scenarios, we provide visual comparisons between our method and four state-of-the-art research models—CatVTON, IDM, StableGarment, and OOTD—on the VITON-HD and DressCode datasets in Figure \ref{fig:5}. 
\begin{figure}[h!] % [h!] 选项建议LaTeX将图片放在代码所在位置 (here)
    \centering % 图片居中
    \includegraphics[width=0.9\textwidth]{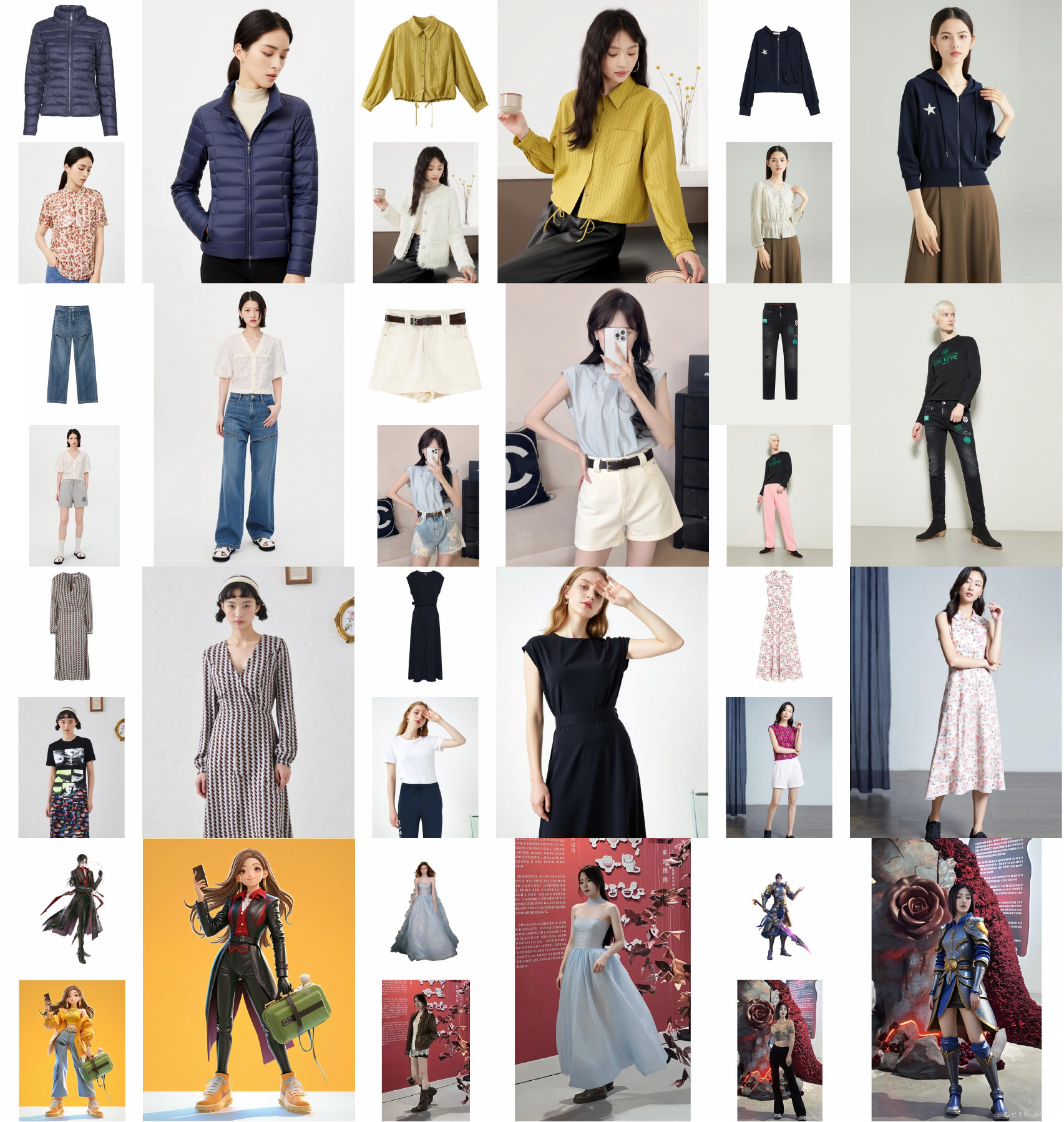}
    \caption{Our model’s results on wild images for upper-body, lower-body, and dress try-on scenarios.} % 添加图片标题
    \label{fig:7} % 添加标签，用于在文中引用
\end{figure}

These comparisons reveal a pronounced advantage in garment consistency, subject clarity, and overall harmony. To further substantiate our results, we conducted rigorous evaluations across multiple try-on tasks under varying conditions. Finally, we benchmarked our model against leading commercial systems (Kling, OutfitAnyone, and GPT-4o) on wild images. The sample outputs in Figure \ref{fig:6} clearly show that our approach achieves more stable performance, adapts effectively to different styles, and consistently yields high-resolution, highly consistent results.Finally, to more effectively demonstrate our model’s performance in real-world tests, we present in Figure \ref{fig:7}a a selection of representative results for intuitive illustration.

%% file: sec/5_conclusion.tex
\section{Discussion and Conclusion}
In this work, we present JCo-MVTON, a novel mask-free virtual try-on model built upon the Multi-Modal Diffusion Transformer (MMDiT) architecture. Inspired by the success of MMDiT in multi-modal generation, we adapt this powerful framework to the multi-conditional nature of virtual try-on, where the generation process must simultaneously respect the reference garment, the wearer’s pose and body structure, and the original image context. Unlike conventional approaches that rely on explicit segmentation masks or dense pose annotations, JCo-MVTON operates directly on RGB inputs, enabling end-to-end training without auxiliary supervision and improving robustness to real-world variations in clothing style and human posture.

To address the scarcity of high-quality, diverse try-on data, we introduce a dual-stream data construction pipeline that leverages generative models to synthesize large-scale, realistic training samples. One stream generates coherent garment-person compositions through deformation and texture transfer, while the other enhances local fidelity and background consistency via diffusion-based inpainting. We further refine the MMDiT backbone by designing an adaptive attention masking mechanism that reduces interference between conditional inputs, along with a pose-aware positional encoding strategy that improves spatial alignment. These architectural enhancements lead to more accurate garment draping and stronger preservation of person-specific details and background content.

Extensive experiments on both consumer-grade and in-the-wild images demonstrate that JCo-MVTON outperforms existing commercial and academic models in visual realism, structural coherence, and identity preservation. While the current implementation shows promising results, we observe limitations in handling fine-grained details—such as hand gestures and small accessories—and occasional instability in challenging poses. Future work will focus on refining detail reconstruction through higher-fidelity data curation and exploring more robust architectural designs for practical deployment in real-world fashion applications.

% The code is publicly available at \textbf{\url{https://github.com/NUS-HPC-AI-Lab/Dynamic-Diffusion-Transformer}}.

%  Building on these designs, we further enhance DyDiT in three key aspects.  First, we reveal that DyDiT integrates seamlessly with flow matching-based generation, enhancing its versatility. Furthermore, we specifically enhance DyDiT to tackle more complex visual generation tasks, including video generation and text-to-image generation, thereby broadening its real-world applications.  Finally, we investigate the feasibility of training DyDiT in a parameter-efficient manner and introduce timestep-based dynamic LoRA (TD-LoRA) to effectively leverage trainable parameter.
% Extensive experiments on diverse visual generation models, including DiT, SiT, Latte, and FLUX, demonstrate the effectiveness of our method. Remarkably, with $<$3\% additional fine-tuning iterations, our approach reduces the FLOPs of DiT-XL by 51\%, accelerates generation by 1.73$\times$, and achieves a competitive FID score of 2.07 on ImageNet.  The code is publicly available at \textbf{\url{https://github.com/NUS-HPC-AI-Lab/Dynamic-Diffusion-Transformer}}.

% \vspace{-2mm}
% \paragraph{Limitations and future works.}
% \noindent
% Similarly to DiT, the proposed DyDiT is currently focusing on image generation. In future works, DyDiT could be further explored to be applied to other tasks, such as video generation~ \cite{ma2024latte} and controllable generation~\cite{chen2024pixart}. 